%% file: main.tex
\documentclass{article}

\usepackage[preprint]{neurips_2023}

\usepackage[utf8]{inputenc} 
\usepackage[T1]{fontenc}    
\usepackage{hyperref}       
\usepackage{url}            
\usepackage{booktabs}       
\usepackage{amsfonts}       
\usepackage{nicefrac}       
\usepackage{microtype}      
\usepackage{xcolor}         

\usepackage{amsmath}
\usepackage{xspace}
\usepackage{amssymb}
\usepackage{graphicx}
\usepackage{multirow}
\usepackage{enumitem}
\usepackage{ulem}
\usepackage{float}
\usepackage{makecell}
\usepackage{caption}
\usepackage{CJKutf8}

\usepackage{fancyhdr}
\newcommand{\modelname}{RoboEgo\xspace}

\title{\modelname System Card:\\ An Omnimodal Model with Native Full Duplexity}

\author{
  Yiqun Yao\textsuperscript{1\textdagger},
  Xiang Li\textsuperscript{1\textdagger}, 
  Xin Jiang\textsuperscript{1\textdagger}, 
  Xuezhi Fang\textsuperscript{1\textdagger}, 
  Naitong Yu\textsuperscript{1\textdagger}, \\
  \textbf{Aixin Sun\textsuperscript{2}, 
  Yequan Wang\textsuperscript{1$*$} }\\
  $^{1}$Beijing Academy of Artificial Intelligence, Beijing, China\\
  $^{2}$School of Computer Science and Engineering, Nanyang Technological University, Singapore
    }

\begin{document}

\begin{figure*}[b]
    \centering
    \includegraphics[scale=0.5]{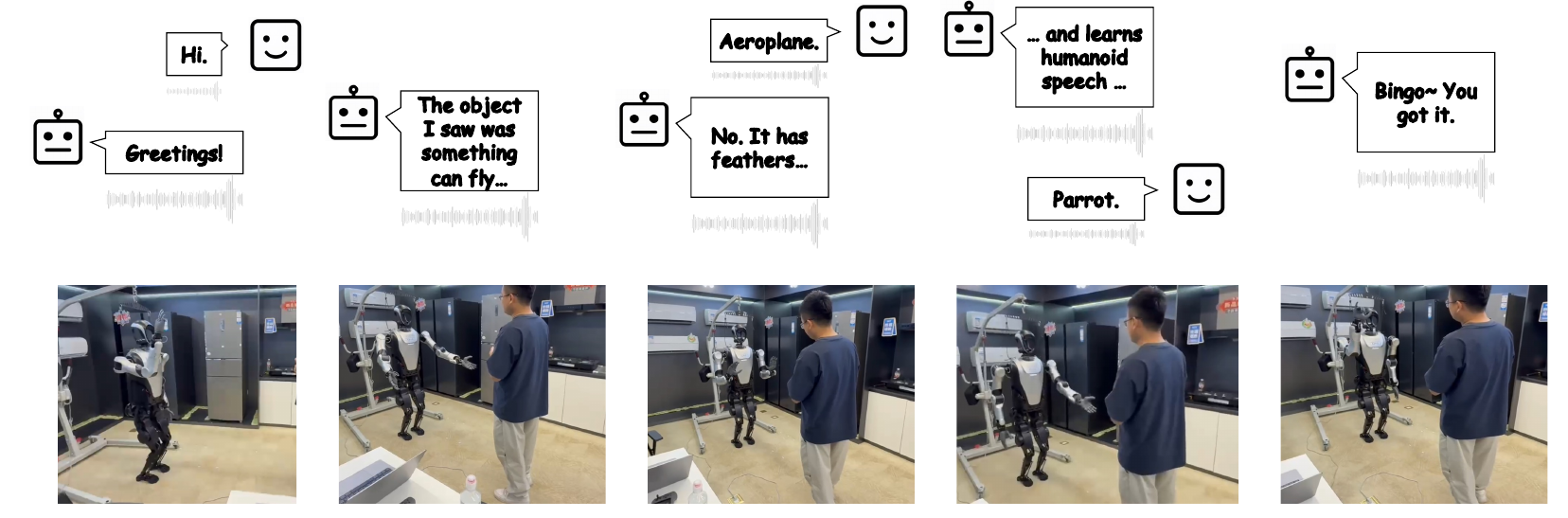}
    \caption{\textbf{The ``Telepathy Challenge'' game} (\begin{CJK}{UTF8}{gbsn}{\scriptsize 心有灵犀}\end{CJK}): a real demo for \modelname's omnimodal full-duplex capabilities.}
    \label{fig:levels}
\end{figure*}

\maketitle

\renewcommand{\thefootnote}{\fnsymbol{footnote}}
\footnotetext[2]{Indicates equal contribution.}
\footnotetext[1]{Corresponding authors.}
\renewcommand{\thefootnote}{\arabic{footnote}}

\begin{abstract}

Humans naturally process real-world multimodal information in a full-duplex manner. In artificial intelligence, replicating this capability is essential for advancing model development and deployment, particularly in embodied contexts. The development of multimodal models faces two primary challenges: (1) effectively handling more than three modalities—such as vision, audio, and text; and (2) delivering full-duplex responses to rapidly evolving human instructions. To facilitate research on models that support both omnimodal processing and full duplexity, we present \modelname (alias: FLM-Ego), a unified model system designed to address both challenges. \modelname incorporates a backbone architecture and algorithms that natively support full duplexity, achieving a theoretical duplex latency of 80 ms. In streaming visually grounded conversations under real-world conditions, \modelname exhibits superior responsiveness and speech naturalness, while maintaining comparable content qualities to state-of-the-art semi-duplex omnimodal models-a feat previously considered unattainable by native full-duplex systems.

\end{abstract}

\section{Introduction}
\label{sec:intro}
The human brain is inherently an \textit{omnimodal} and \textit{full-duplex} system. Omnimodal capability refers to the ability to process and generate information across all modalities, including text, audio, images, video, 3D data, and embodied actions, among other sensory inputs. Full duplexity denotes the parallel execution of input, processing, and output across modalities, enabling real-time responsiveness to dynamic environments. In the context of artificial general intelligence (AGI), omnimodal capability is widely regarded as a fundamental and long-term objective \cite{sparks, how_far_agi}, while full duplexity is increasingly valued for its lower overhead and improved interaction experience \cite{beyond,full-dup}. In the embodied intelligence literature \cite{embodied}, both capabilities are recognized as critical requirements for achieving L3+ embodied AGI \cite{towardembodied}. Specifically, (1) a robotic brain with strong generalization ability must possess comprehensive understanding of the physical world, which is largely mediated by its modality coverage; and (2) native duplexity enables real-time responsiveness, essential for advanced humanoid applications.

Large language models (LLMs) \cite{GPT-4, deepseek-v3} have rapidly advanced and are being extended to audio-language models \cite{audiolm, glm-voice, mini-omni, qwen-audio, moshi}, visual-language models (VLMs) \cite{llava, pixtral, cambrian}, and visual-language-action (VLA) models \cite{RT-2, openvla}. However, these models often lack either visual or auditory modality support, limiting their versatility. Recent progress in omnimodal models—capable of simultaneously handling text, audio, and visual modalities—has attracted significant attention \cite{4o, gemini1.5, qwen-omni, minicpm-o}. Yet, the integration of full duplexity into omnimodal systems remains underexplored. Most current approaches rely on time-division multiplexing (TDM) strategies \cite{beyond, full-dup, omniflatten, minmo}. Since the widely adopted Transformer attention mechanism \cite{vaswani2017attention} has a computational complexity of $O(n^2)$, extending its input via TDM (e.g., interleaving listening and speaking tokens in a single stream) significantly hampers responsiveness—resulting in full-duplex delays of up to 2 seconds \cite{beyond}—and limits generation length, with audio capped at around 45 seconds \cite{minicpm-o}. This limitation becomes increasingly restrictive as omnimodal foundation models continue to \textit{scale up} \cite{chinchilla, mu-scaling, o1}.

To address these challenges, we introduce \modelname, an omnimodal model capable of handling a wide spectrum of tasks across text, vision, audio, and embodied interaction, while natively supporting full duplexity in multiple modalities:  \textit{\modelname watches, thinks, listens, speaks, and acts simultaneously}, with a theoretical response granularity of 80 ms for tasks such as user instruction following, interruption handling, and turn-taking.

\modelname’s training framework introduces innovations in the organization of full-duplex omnimodal data streams,training stage arrangement, and nuanced objective designs. These improvements reduce modality conflicts, enhance scalability to more modalities, and boost robustness in real-world applications.

We evaluate \modelname on a broad set of omnimodal tasks, including visual understanding, speech comprehension and generation, embodied actions, and real-time multimodal dialogue. Both automatic metrics and human evaluations demonstrate that \modelname achieves state-of-the-art performance in conversational quality across multiple skill dimensions, while maintaining competitive results on standard benchmarks against strong modality-specific models.

\section{Model Architecture}
\label{sec:architecture}

\begin{figure*}
    \centering
    \includegraphics[scale=0.48]{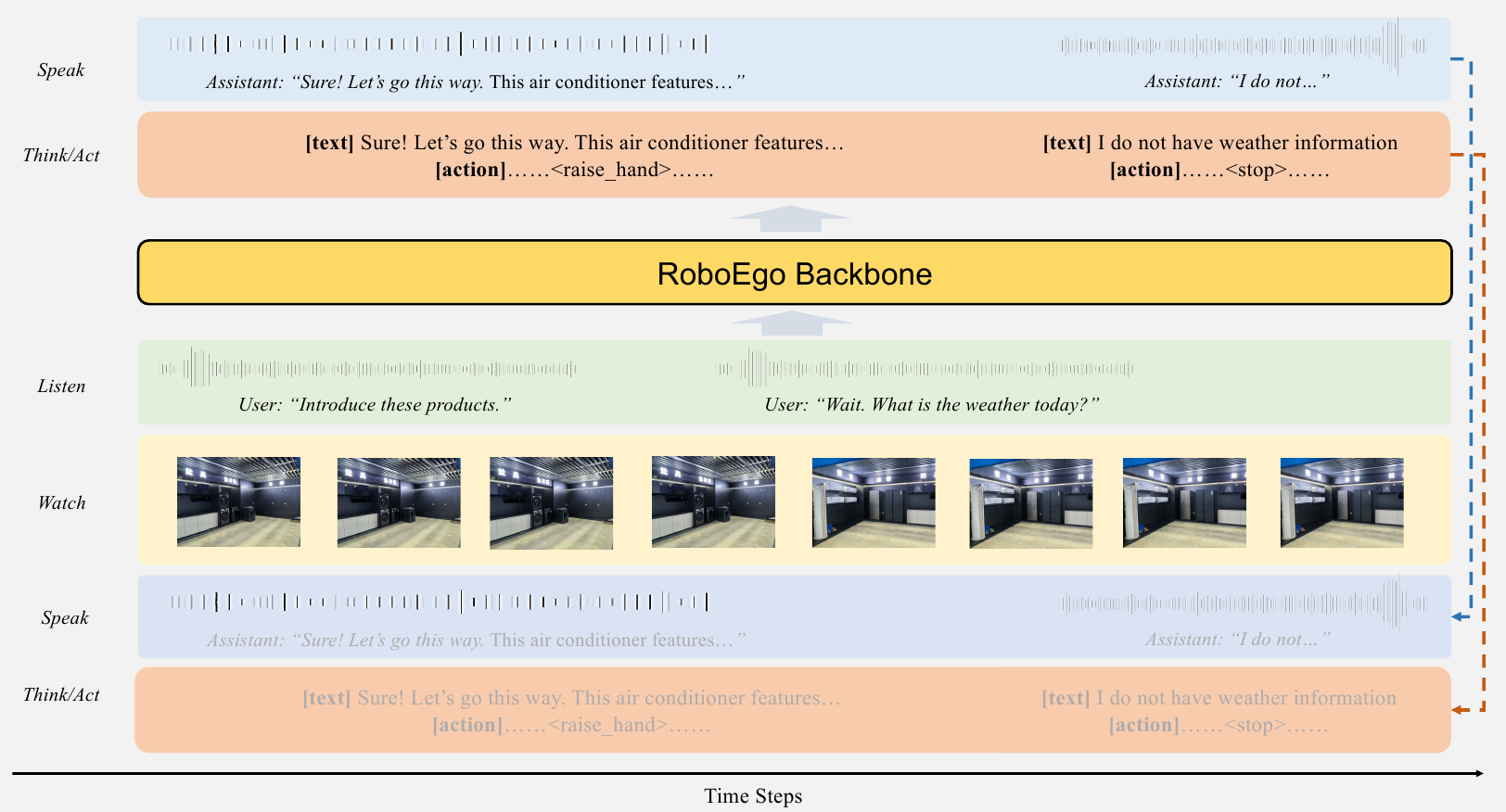}
    \caption{\textbf{Illustration of the architecture of \modelname}.}
    \label{fig:architecture}
\end{figure*}

The framework of \modelname is illustrated in Figure~\ref{fig:architecture}. To ensure inherent responsiveness, the model is designed to receive real-time streaming input while simultaneously generating textual responses, audio outputs, and embodied actions. At each time step of the backbone model, input streams are efficiently merged, e.g., combining real-time listening input with previously generated speech, text, and action outputs. This parallel processing approach eliminates the need for time-slice sharing via time-division multiplexing (TDM), which is commonly employed in prior work such as MiniCPM-o~\cite{minicpm-o} and Qwen2.5-Omni~\cite{qwen-omni}. The visual stream can be flexibly utilized either as contextual information through TDM—leveraging cross-attention in the backbone—or integrated directly into the parallel streams, depending on the specific task. When stream-level visual embeddings are employed, \modelname realizes a complete prototype instantiation of a model architecture that satisfies the requirements for L3+ embodied AI~\cite{towardembodied}.

\subsection{Backbone Structure}

Our backbone model is a 7-billion-parameter autoregressive LLM~\cite{llama3, qwen3}. At each time step, the model processes the following inputs: contextual visual embeddings $v_{1...m}^{context}$ or time-aligned visual embeddings $v_{t}^{stream}$ (depending on the task), all prior audio tokens from both the speaking and listening channels $a_{\sim t-1}^{speak},a_{\sim t-1}^{listen}$, and all preceding textual tokens $w_{\sim t-1}^{text}$. These inputs are fused to produce informative hidden states, which are subsequently used for decoding textual and audio outputs:

\begin{equation}
    h_t = F_{\theta}^{back}(v_{1...m}^{context},a_{\sim t-1}^{speak},a_{\sim t-1}^{listen},w_{\sim t-1}^{text}).
    \label{eq:backbone}
\end{equation}

\subsection{Modality-specific Modules}
We apply visual encoders, e.g., Vision Transformers \cite{vit,qwen2.5vl}, to generate contextual or streaming visual embeddings, which summarizes relevant information from image sequences and videos:

\begin{align}
    v_{1...m}^{context} &= Enc_v(i_1...i_n), or\\
    v_{t}^{stream} &= Enc_v(i_{\sim t-1}).
\end{align}

We apply discrete audio codecs to represent and generate both English and Chinese speech in full-duplex speaking and listening streams. Textual tokens are handled by standard LLM tokenizers.

At each time step, lightweight decoders for audio, language, and action modalities generate their respective tokens based on a shared hidden state produced by the backbone model. This unified decoding process ensures synchronized, modality-consistent outputs across speech, text, and embodied actions:

\begin{align}
    id_t^{aud/txt/act} &= Dec_{aud/txt/act}(h_t),
\end{align}

We observe that the hidden state $h_t$ is sufficiently informative to support simultaneous generation across audio, text, and action modalities. As a result, our decoders operate locally at each time step $t$, without the need to re-aggregate $O(N^2)$ contextual information—unlike the "talker-like" architectures employed in related work~\cite{qwen-omni}.

\subsection{Stream Organization}

\begin{figure*}
    \centering
    \includegraphics[scale=0.57]{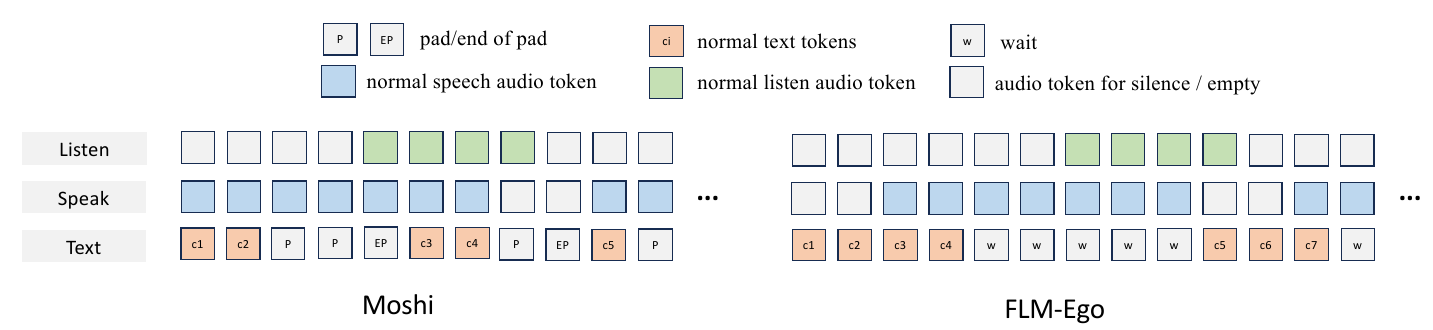}
    \caption{\textbf{Stream organization for text and audio in \modelname}.}
    \label{fig:duplex}
\end{figure*}

Even for a single aligned utterance, textual and audio tokens are naturally asynchronous: one second of speech—typically represented by over 10 frames of audio features—often corresponds to only 3–4 textual tokens. To address this mismatch, Moshi~\cite{moshi} introduces a token-level alignment strategy, where textual tokens are split using special "pad" and "end-of-pad" tokens, allowing each token to appear precisely at the time it is spoken (Figure~\ref{fig:duplex}, left). However, this approach has two key limitations: (1) it requires fine-grained, word-level timestamps for training annotations, which significantly increases data processing cost and introduces vulnerability to cascading alignment errors; and (2) it deviates from the humanoid pattern of monologue generation—in real-world scenarios, humans think, listen, and speak concurrently, with internal thoughts generally forming a coherent, forward-progressing stream that precedes speech. From a machine learning perspective, breaking sentences into disjointed word-level fragments also degrades language modeling performance, as noted in the original work.

To overcome these limitations, \modelname adopts a "text-first" strategy, illustrated in Figure~\ref{fig:duplex} (right). For a given utterance beginning at time step $t$, we place its corresponding semantic token at $t$, and initiate the full textual monologue at $t-spk\_delay$. This $spk\_delay$ mimics human-like cognitive processes—such as thinking, planning, and phrase organization—that occur prior to speech. Once the textual monologue ends, the remaining time in the text channel is filled with a special \texttt{<wait>} token until the corresponding speech output completes. This strategy requires only sentence-level transcriptions for training, substantially reducing annotation effort and enabling large-scale data scaling. Furthermore, it preserves the autoregressive language modeling capabilities of the pretrained backbone.

\section{Training Paradigm}
\label{sec:training}
Although training from scratch across all modalities is feasible, we initialize the \modelname backbone using pre-trained language models and visual encoders prior to training. Audio embeddings and associated modules are randomly initialized. This initialization strategy significantly reduces computational cost while remaining effective for validating the core concepts of omnimodality and full duplexity. The training process of \modelname consists of two stages: post-training and fine-tuning.

\subsection{Post-training}

In the post-training stage, \modelname is equipped with foundational unimodal and cross-modal capabilities necessary for real-time omnimodal applications. We begin by introducing audio-oriented capabilities to the backbone model using a large-scale corpus of audio data, while preserving the language modeling and visual-language alignment abilities of the pre-trained foundation model. This stage encompasses a broad spectrum of tasks, including automatic speech recognition (ASR), text-to-speech synthesis (TTS), visual question answering (VQA), optical character recognition (OCR), and more.

\subsection{Supervised Fine-tuning (SFT)}
In this stage, \modelname is fine-tuned to function as a general-purpose omnimodal chatbot. The objective is to develop capabilities for speech-based instruction following, visually grounded spoken dialogue, full-duplex turn-taking, and accurate action execution in embodied environments. To this end, we primarily utilize multi-turn, visually grounded speech dialogues—both with and without associated actions—as the core training data. This dataset is further augmented to support full-duplex interruption handling and to enhance robustness against environmental noise.

Notably, because full duplexity is intrinsically supported by the \modelname architecture, the model learns to respond in a full-duplex manner through standard supervised learning with appropriate training data, without requiring complex engineering interventions or specialized training mechanisms.

\section{Experiments}
\label{sec:results}
The ultimate goal of \modelname is to develop an omnimodal assistant equipped with native full-duplex capabilities. To evaluate the maturity of \modelname in achieving this objective, we focus on the following research questions:
\begin{itemize}
\item \textbf{RQ1:} Has \modelname achieved performance comparable to specialized models on relevant unimodal or bimodal tasks?
\item \textbf{RQ2:} How does the model’s joint-modal capability evolve across different training stages?
\item \textbf{RQ3:} Does the inherent full-duplex architecture of \modelname lead to improved user experience in omnimodal applications?
\end{itemize}

To address RQ1, we compare \modelname against both specialized models and multimodal models across a range of subtasks, including visual understanding, and audio understanding/generation. We highlight key architectural differences and analyze their potential impact on performance. For RQ2, we track the evolution of each multimodal capability throughout the training process, with a focus on how these capabilities are introduced, integrated, and preserved across stages. To answer RQ3, we perform both automatic and human evaluations on real-world omnimodal tasks, assessing performance across multiple dimensions of user experience and interaction quality.

\subsection{Visual Understanding}

We evaluate the visual understanding tasks by generating answers in the text channel. We choose MMStar \cite{mmstar}, RealWorldQA \cite{realworldqa}, and OCRBench \cite{ocrbench} to test the model's image understanding abilities. For video comprehension, we directly compare the ultimate omnimodal-chatting in Section \ref{sec:omni-chatting}.

For specialized visual-language model, we choose Qwen2.5-VL \cite{qwen2.5vl} as baseline. For proprietary models, we compare to GPT-4o-mini, GPT-4o \cite{4o}, and Gemini-1.5-Pro \cite{gemini1.5}. For other omnimodal baselines, we list Mini-CPM-o \cite{minicpm-o} and Qwen2.5-omni \cite{qwen-omni} as references.

We present the visual understanding results in Table~\ref{tab:visual_results}. The results show that \modelname achieves performance comparable to or better than proprietary models such as GPT-4o-mini and Gemini-1.5-Pro on visual understanding benchmarks. Notably, it matches the performance of state-of-the-art specialized models—such as Qwen2.5-VL-7B—on the RealWorldQA task. By comparing results from the post-training and SFT stages, we observe that the supervised fine-tuning process effectively adapts the model's capabilities from text-based VQA to speech-based VQA, with only minimal performance degradation.
\input{tables/visual_understanding}

\subsection{Audio Understanding and Generation}
We use word error rate (WER) as the primary metric for evaluating audio understanding in automatic speech recognition (ASR). Evaluations are conducted on both Chinese and English ASR benchmarks, including Fleurs-zh~\cite{fleurs}, LibriSpeech-clean~\cite{librispeech}, and WenetSpeech-net~\cite{wenetspeech}. Although audio instruction-following is evaluated separately in Section~\ref{sec:omni-chatting}, we also include LlamaQuestions~\cite{llamaquestions} as an audio-based question answering benchmark in this section.

Among the compared methods, Whisper-large-v3~\cite{whisper}, Qwen2-Audio~\cite{qwen-audio}, MinMo~\cite{minmo}, and GLM-4-Voice~\cite{glm-voice} are specialized bimodal audio-language models. Omnimodal baselines include GPT-4o~\cite{4o}, Gemini 1.5 Pro~\cite{gemini1.5}, MiniCPM-o~\cite{minicpm-o}, and Qwen2.5-Omni~\cite{qwen-omni}. We also compare to Moshi~\cite{moshi}, a native full-duplex audio-language model, as a key baseline for duplex-capable systems.

The results for ASR and speech-based question answering are presented in Table~\ref{tab:audio_understanding}. Following both post-training and supervised fine-tuning (SFT), \modelname demonstrates competitive performance on Chinese ASR, outperforming specialized models such as Qwen2-Audio on the Fleurs evaluation set. In the LibriSpeech-clean benchmark, \modelname surpasses Moshi, an English-specialized full-duplex system. Notably, ASR performance improves after SFT in both benchmarks, attributed to the increased inclusion of speech data during fine-tuning. However, on the WenetSpeech-net dataset, we observe a performance drop after SFT. We hypothesize that this degradation is due to the dataset’s increased complexity in dialects. On LlamaQuestions, \modelname also reaches the performance level of related Chinese-English models, indicating that its textual knowledge capabilities are well preserved throughout the training process.

\input{tables/audio_understanding}

We evaluate audio generation capabilities on the Seed-TTS-en and Seed-TTS-zh benchmarks~\cite{seed-eval}, following standard evaluation protocols. The results are summarized in Table~\ref{tab:audio_generation}. \modelname achieves WER scores comparable to those of advanced, specialized text-to-speech (TTS) models. Although \modelname is not designed to prioritize high-fidelity voice cloning, lightweight fine-tuning enables it to produce speech with greater audio similarity than other omnimodal systems including MiniCPM-o.

\input{tables/audio_generation}

\subsection{Real-time Omnimodal Chatting}
\label{sec:omni-chatting}

Real-time omnimodal chatting represents one of the ultimate design goals of \modelname. In this setting, the AI assistant engages in duplex conversations with a human user while simultaneously processing image or video inputs—either uploaded by the user or streamed via webcam. This task differs significantly from traditional textual multimodal instruction-following, particularly in terms of human preference. In instruction-following, users often favor detailed textual responses, especially for tasks involving programming or complex reasoning~\cite{o1, dpskr1}. In contrast, during natural conversations, users tend to prefer concise, summarized, or even intentionally evasive answers.

To support a comprehensive, multi-dimensional evaluation, we report both automatic metrics and human-annotated scores. For automatic evaluation, we construct a speech instruction-following test set using publicly available Chinese prompts formatted similarly to AlpacaEval~\cite{alpaca_eval}. We employ DeepSeek-V3~\cite{deepseek-v3} as a reference model to produce quality scores (ranging from 0 to 10) by comparing model outputs to ground truth responses. For human evaluation, we conduct a double-blind head-to-head comparison between \modelname and Qwen2.5-Omni~\cite{qwen-omni}, a state-of-the-art streaming chatbot. Five human annotators rate multi-turn audio responses across four dimensions: (1) Helpfulness, measuring the informativeness and relevance of the content; (2) Naturalness, evaluating alignment with conversational tone and linguistic style; (3) Responsiveness, assessing reaction speed to interruptions and dynamic input; and (4) Robustness, measuring stability under diverse real-world noise conditions.

The automatic scores and human-annotated scores are summarized in Table~\ref{tab:omni_chatting}. Compared to the advanced omnimodal chatting system Qwen2.5-omni, \modelname delivers responses of comparable quality in terms of helpfulness, as confirmed by both automated and human assessments. Moreover, in metrics specific to real-time conversational scenarios—naturalness, responsiveness, and robustness to noise—\modelname exhibits a clear advantage. We attribute this performance to the model’s inherent full-duplex architecture and the effectiveness of the training paradigm.

\input{tables/omni_chatting}

\subsection{Embodied Actions}

To evaluate real-time embodied action-taking, we deploy \modelname on LEJU Kuavo robots \footnote{\url{https://www.lejurobot.com/en}} and assess its performance on two embodied tasks: Locomotion and the Telepathy Challenge. In the locomotion task, the model is required to interpret speech instructions and visual input from the robot’s camera feed to generate appropriate movement actions, resulting in correct spatial mobility of the robot. In the Telepathy Challenge, the robot participates in a collaborative game where it is shown an image that is hidden from its human teammate. The robot must describe the object using speech—without explicitly naming it—to help the human identify it. Additionally, \modelname generates gesture sequences based on the human’s spoken responses to convey emotions and provide subtle clues about the object and game progress.

The success rates for these tasks are reported in Table~\ref{tab:embodied_action}. The results show that \modelname achieves high accuracy in generating contextually appropriate actions based on real-time, multimodal input—including visual streams, duplex audio, and text. To our knowledge, this is the first demonstration of such embodied action-taking capabilities by an omnimodal model.

\input{tables/embodied_action}

\subsection{Answers to Research Questions}
To summarize the experimental analysis, we revisit and answer the research questions posed at the beginning:

\textit{RQ1: Has \modelname reached the performance level of specialized models in relevant unimodal or bimodal tasks?}
In audio-centric tasks such as ASR and TTS, \modelname demonstrates competitive performance, comparable to state-of-the-art specialized and omnimodal models. This establishes a strong foundation for real-time duplex audio interaction. In speech and visual understanding, \modelname achieves approximately 90\% of the performance of leading open-sourced or proprietary systems, enabling reliable perception of visual environments in most everyday scenarios.

\textit{RQ2: How does the model’s joint-modal capability evolve across training stages?}
The visual understanding performance is established in the pre-training and post-training process and maintained with minimal degradation in subsequent phases. For the audio modality, both ASR and TTS capabilities are developed from scratch during post-training and are progressively refined across stages. Following supervised fine-tuning (SFT), \modelname acquires specialized competencies for real-time omnimodal dialogue and embodied action-taking.

\textit{RQ3: Does the key advantage of inherent duplexity translate into improved user experience in omnimodal applications?}
As demonstrated in Section~\ref{sec:omni-chatting}, the answer is affirmative. The results indicate that native full-duplexity not only enhances responsiveness and naturalness in interaction but also represents a practical and effective design paradigm for future omnimodal AI systems.

\section{Conclusion and Future Challenges}
In this paper, we introduce \modelname, an omnimodal system with native full-duplex capabilities—\modelname watches, thinks, listens, speaks, and acts simultaneously. We envision \modelname as an initial step toward exploring architectural foundations that meet the demands of broad modality coverage and real-time responsiveness required for L3+ embodied AGI. Compared to existing systems, \modelname (FLM-Ego) demonstrates notable advantages in delivering smoother experiences in streaming omnimodal dialogue and supporting embodied actions, while maintaining competitive performance on standard benchmarks.

Despite these advancements, challenges remain. For certain visually intensive tasks requiring fine-grained visual understanding, current stream-based visual encoders still fall short in reliability for full-duplex, all-modality processing, necessitating the use of time-division multiplexing (TDM) as a temporary workaround. Future breakthroughs in long-context visual understanding will be critical to advancing \modelname-like systems. Additional challenges include expanding modality coverage (e.g., tactile perception, 3D spatial representations), resolving cross-modal conflicts, and enabling more precise and nuanced robotic actions.

\section*{Acknowledgments}
We would like to thank the colleagues from Beijing Academy of Artificial Intelligence (BAAI) and LEJU Robot for their help on computational resources and experimental devices, and all other colleagues' strong support for this project.

\bibliographystyle{plain}
\bibliography{custom}

\end{document}

%% file: tables/visual_understanding.tex
\begin{table*}[thbp]
\centering
\caption{\textbf{Visual understanding results.} We summarize the benchmark results in each training stage, compared to strong baselines.}
\scalebox{0.88}
{
\begin{tabular}{c|ccc}
\toprule
\multirow{2}{*}{Model} & \multicolumn{3}{c}{Image Understanding}\\
 & MMStar   & RealWorldQA & OCRBench  \\
\midrule
Qwen2.5vl-7B & 63.9 & 68.5 & 864  \\\midrule
GPT-4o-mini & 54.8 & 23.5 & -  \\
GPT-4o & 63.9& 47.6 & 736 \\
Gemini-1.5-pro & 59.1& 51.5 &754 \\\midrule
MiniCPM-o & 64.0 & - & 897 \\
Qwen-2.5-omni & 64.0& 70.3 & - \\
\midrule
\modelname (Post) & 53.8 & 66.4 & 772  \\
\modelname (SFT) & 50.0 & 63.1& 770 \\
\bottomrule
\end{tabular}
}
\label{tab:visual_results}
\end{table*}

%% file: tables/audio_understanding.tex
\begin{table*}[thbp]
\centering
\caption{\textbf{Audio understanding results.} We include ASR and audio question answering benchmarks.}
    \begin{tabular}{l|ccc|c}
        \toprule
        \multirow{2}{*}{Model} & Fleurs & WenetSpeech & LibriSpeech & \multirow{2}{*}{LlamaQuestions}  \\
        & zh & net & clean & \\
        \midrule
        Whisper-large-v3 & 7.7 & - & 1.8 & -  \\
        Qwen2-Audio      & 7.5   & - & 1.6 & - \\
        MinMo            & 3.0 & 6.8 & 1.7 & 64.1 \\
        GLM-4-Voice      & - & - & 2.8 & 50(64.7) \\\midrule
        GPT-4o           & 5.4   & - & -  & 71.7 \\
        Gemini-1.5-pro   & 5.9   & 14.3 & - & \\
        MiniCPM-o        & 4.4 & 6.9 & 1.7 & 61 \\
        Qwen2.5-omni     & 3.0 & 5.9 & 1.8 & - \\\midrule
        Moshi            & - & - & 5.7 & 43.7(62.3) \\\midrule
        \modelname (Post)     & 5.5  & 16.2 & 4.6 & - \\
        \modelname (SFT) & 5.4 & 17.1 & 3.2 & 56.3  \\
        \bottomrule
    \end{tabular}
    \label{tab:audio_understanding}
\end{table*}

%% file: tables/audio_generation.tex

\begin{table*}[thbp]
\caption{\textbf{Audio generation results.} We include WER and speaker similarity as metrics.}
\centering
    \begin{tabular}{l|cccc}
        \toprule
        \multirow{2}{*}{Model} & \multicolumn{2}{c}{Seed-tts-en} & \multicolumn{2}{c}{Seed-tts-zh} \\
        & WER & SIM & WER & SIM \\
        \midrule
        Seed-tts      & 2.25 & 0.762 & 1.12 & 0.796 \\
        Cosyvoice     & 4.29 & 0.609 & 3.63 & 0.723 \\
        Cosyvoice2    & 2.57 & 0.652 & 1.45 & 0.748 \\\midrule
        GLM-4-VOICE   & 2.91 & - & 2.10 & - \\
        Minmo         & 2.90 & - & 2.48 & - \\\midrule
        MiniCPM-o     & - & 0.470 & -  & 0.570 \\
        Qwen-2.5-omni & 2.72 & 0.632 &1.70 & 0.752 \\
        \midrule
        \modelname (SFT)   & 2.95     &   0.543   &    2.10   &    0.601    \\
        \bottomrule
    \end{tabular}
    \label{tab:audio_generation}
\end{table*}

%% file: tables/omni_chatting.tex
\begin{table*}[thbp]
\centering
\caption{\textbf{Omnimodal chatting results.} Automatic and human evaluation results are included.}
\scalebox{0.75}
{
\centering
    \begin{tabular}{l|c|cccc}
        \toprule
        \multirow{2}{*}{Model} & Instruct & \multicolumn{4}{c}{Visual Grounded} \\
       & LLM-score  & Helpfulness & Naturalness & Responsiveness & Robustness \\
        \midrule
        Qwen-2.5-omni & 6.36 & 7.4 & 7.9 & 8.1 & 7.7 \\
        \midrule
        \modelname       & 6.58 & 7.2 & 8.2 & 8.8 &8.0   \\
        \bottomrule
    \end{tabular}
    \label{tab:omni_chatting}
}
\end{table*}

%% file: tables/embodied_action.tex
\begin{table*}[thbp]
\centering
\caption{\textbf{Accuracy for embodied action tasks.}}
\scalebox{0.95}
{
\centering
    \begin{tabular}{l|c|c}
        \toprule
        Model & Locomotion & ``Telepathy Challenge'' \\
        \midrule
        \modelname (SFT)       & 96.5\% & 97.2\%   \\
        \bottomrule
    \end{tabular}
    \label{tab:embodied_action}
}
\end{table*}

%% file: main.bbl
\begin{thebibliography}{10}

\bibitem{pixtral}
Pravesh Agrawal, Szymon Antoniak, Emma~Bou Hanna, Baptiste Bout, Devendra~Singh Chaplot, Jessica Chudnovsky, Diogo Costa, Baudouin~De Monicault, Saurabh Garg, Th{\'{e}}ophile Gervet, Soham Ghosh, Am{\'{e}}lie H{\'{e}}liou, Paul Jacob, Albert~Q. Jiang, Kartik Khandelwal, Timoth{\'{e}}e Lacroix, Guillaume Lample, Diego de~Las~Casas, Thibaut Lavril, Teven~Le Scao, Andy Lo, William Marshall, Louis Martin, Arthur Mensch, Pavankumar Muddireddy, Valera Nemychnikova, Marie Pellat, Patrick von Platen, Nikhil Raghuraman, Baptiste Rozi{\`{e}}re, Alexandre Sablayrolles, Lucile Saulnier, Romain Sauvestre, Wendy Shang, Roman Soletskyi, Lawrence Stewart, Pierre Stock, Joachim Studnia, Sandeep Subramanian, Sagar Vaze, Thomas Wang, and Sophia Yang.
\newblock Pixtral 12b.
\newblock {\em CoRR}, abs/2410.07073, 2024.

\bibitem{seed-eval}
Philip Anastassiou, Jiawei Chen, Jitong Chen, Yuanzhe Chen, Zhuo Chen, Ziyi Chen, Jian Cong, Lelai Deng, Chuang Ding, Lu~Gao, Mingqing Gong, Peisong Huang, Qingqing Huang, Zhiying Huang, Yuanyuan Huo, Dongya Jia, Chumin Li, Feiya Li, Hui Li, Jiaxin Li, Xiaoyang Li, Xingxing Li, Lin Liu, Shouda Liu, Sichao Liu, Xudong Liu, Yuchen Liu, Zhengxi Liu, Lu~Lu, Junjie Pan, Xin Wang, Yuping Wang, Yuxuan Wang, Zhen Wei, Jian Wu, Chao Yao, Yifeng Yang, Yuanhao Yi, Junteng Zhang, Qidi Zhang, Shuo Zhang, Wenjie Zhang, Yang Zhang, Zilin Zhao, Dejian Zhong, and Xiaobin Zhuang.
\newblock Seed-tts: {A} family of high-quality versatile speech generation models.
\newblock {\em CoRR}, abs/2406.02430, 2024.

\bibitem{qwen2.5vl}
Shuai Bai, Keqin Chen, Xuejing Liu, Jialin Wang, Wenbin Ge, Sibo Song, Kai Dang, Peng Wang, Shijie Wang, Jun Tang, et~al.
\newblock Qwen2. 5-vl technical report.
\newblock {\em arXiv preprint arXiv:2502.13923}, 2025.

\bibitem{audiolm}
Zal{\'a}n Borsos, Rapha{\"e}l Marinier, Damien Vincent, Eugene Kharitonov, Olivier Pietquin, Matt Sharifi, Dominik Roblek, Olivier Teboul, David Grangier, Marco Tagliasacchi, et~al.
\newblock Audiolm: a language modeling approach to audio generation.
\newblock {\em IEEE/ACM transactions on audio, speech, and language processing}, 31:2523--2533, 2023.

\bibitem{RT-2}
Anthony Brohan, Noah Brown, Justice Carbajal, Yevgen Chebotar, Xi~Chen, Krzysztof Choromanski, Tianli Ding, Danny Driess, Avinava Dubey, Chelsea Finn, Pete Florence, Chuyuan Fu, Montse~Gonzalez Arenas, Keerthana Gopalakrishnan, Kehang Han, Karol Hausman, Alexander Herzog, Jasmine Hsu, Brian Ichter, Alex Irpan, Nikhil~J. Joshi, Ryan Julian, Dmitry Kalashnikov, Yuheng Kuang, Isabel Leal, Lisa Lee, Tsang{-}Wei~Edward Lee, Sergey Levine, Yao Lu, Henryk Michalewski, Igor Mordatch, Karl Pertsch, Kanishka Rao, Krista Reymann, Michael~S. Ryoo, Grecia Salazar, Pannag Sanketi, Pierre Sermanet, Jaspiar Singh, Anikait Singh, Radu Soricut, Huong~T. Tran, Vincent Vanhoucke, Quan Vuong, Ayzaan Wahid, Stefan Welker, Paul Wohlhart, Jialin Wu, Fei Xia, Ted Xiao, Peng Xu, Sichun Xu, Tianhe Yu, and Brianna Zitkovich.
\newblock {RT-2:} vision-language-action models transfer web knowledge to robotic control.
\newblock {\em CoRR}, abs/2307.15818, 2023.

\bibitem{sparks}
S{\'{e}}bastien Bubeck, Varun Chandrasekaran, Ronen Eldan, Johannes Gehrke, Eric Horvitz, Ece Kamar, Peter Lee, Yin~Tat Lee, Yuanzhi Li, Scott~M. Lundberg, Harsha Nori, Hamid Palangi, Marco~T{\'{u}}lio Ribeiro, and Yi~Zhang.
\newblock Sparks of artificial general intelligence: Early experiments with {GPT-4}.
\newblock {\em CoRR}, abs/2303.12712, 2023.

\bibitem{mmstar}
Lin Chen, Jinsong Li, Xiaoyi Dong, Pan Zhang, Yuhang Zang, Zehui Chen, Haodong Duan, Jiaqi Wang, Yu~Qiao, Dahua Lin, et~al.
\newblock Are we on the right way for evaluating large vision-language models?
\newblock {\em arXiv preprint arXiv:2403.20330}, 2024.

\bibitem{minmo}
Qian Chen, Yafeng Chen, Yanni Chen, Mengzhe Chen, Yingda Chen, Chong Deng, Zhihao Du, Ruize Gao, Changfeng Gao, Zhifu Gao, et~al.
\newblock Minmo: A multimodal large language model for seamless voice interaction.
\newblock {\em arXiv preprint arXiv:2501.06282}, 2025.

\bibitem{qwen-audio}
Yunfei Chu, Jin Xu, Xiaohuan Zhou, Qian Yang, Shiliang Zhang, Zhijie Yan, Chang Zhou, and Jingren Zhou.
\newblock Qwen-audio: Advancing universal audio understanding via unified large-scale audio-language models.
\newblock {\em arXiv preprint arXiv:2311.07919}, 2023.

\bibitem{fleurs}
Alexis Conneau, Min Ma, Simran Khanuja, Yu~Zhang, Vera Axelrod, Siddharth Dalmia, Jason Riesa, Clara Rivera, and Ankur Bapna.
\newblock Fleurs: Few-shot learning evaluation of universal representations of speech.
\newblock {\em arXiv preprint arXiv:2205.12446}, 2022.

\bibitem{dpskr1}
DeepSeek{-}AI, Daya Guo, Dejian Yang, Haowei Zhang, Junxiao Song, Ruoyu Zhang, Runxin Xu, Qihao Zhu, Shirong Ma, Peiyi Wang, Xiao Bi, Xiaokang Zhang, Xingkai Yu, Yu~Wu, Z.~F. Wu, Zhibin Gou, Zhihong Shao, Zhuoshu Li, Ziyi Gao, Aixin Liu, Bing Xue, Bingxuan Wang, Bochao Wu, Bei Feng, Chengda Lu, Chenggang Zhao, Chengqi Deng, Chenyu Zhang, Chong Ruan, Damai Dai, Deli Chen, Dongjie Ji, Erhang Li, Fangyun Lin, Fucong Dai, Fuli Luo, Guangbo Hao, Guanting Chen, Guowei Li, H.~Zhang, Han Bao, Hanwei Xu, Haocheng Wang, Honghui Ding, Huajian Xin, Huazuo Gao, Hui Qu, Hui Li, Jianzhong Guo, Jiashi Li, Jiawei Wang, Jingchang Chen, Jingyang Yuan, Junjie Qiu, Junlong Li, J.~L. Cai, Jiaqi Ni, Jian Liang, Jin Chen, Kai Dong, Kai Hu, Kaige Gao, Kang Guan, Kexin Huang, Kuai Yu, Lean Wang, Lecong Zhang, Liang Zhao, Litong Wang, Liyue Zhang, Lei Xu, Leyi Xia, Mingchuan Zhang, Minghua Zhang, Minghui Tang, Meng Li, Miaojun Wang, Mingming Li, Ning Tian, Panpan Huang, Peng Zhang, Qiancheng Wang, Qinyu Chen, Qiushi Du, Ruiqi Ge,
  Ruisong Zhang, Ruizhe Pan, Runji Wang, R.~J. Chen, R.~L. Jin, Ruyi Chen, Shanghao Lu, Shangyan Zhou, Shanhuang Chen, Shengfeng Ye, Shiyu Wang, Shuiping Yu, Shunfeng Zhou, Shuting Pan, and S.~S. Li.
\newblock Deepseek-r1: Incentivizing reasoning capability in llms via reinforcement learning.
\newblock {\em CoRR}, abs/2501.12948, 2025.

\bibitem{moshi}
Alexandre D{\'{e}}fossez, Laurent Mazar{\'{e}}, Manu Orsini, Am{\'{e}}lie Royer, Patrick P{\'{e}}rez, Herv{\'{e}} J{\'{e}}gou, Edouard Grave, and Neil Zeghidour.
\newblock Moshi: a speech-text foundation model for real-time dialogue.
\newblock {\em CoRR}, abs/2410.00037, 2024.

\bibitem{vit}
Alexey Dosovitskiy, Lucas Beyer, Alexander Kolesnikov, Dirk Weissenborn, Xiaohua Zhai, Thomas Unterthiner, Mostafa Dehghani, Matthias Minderer, Georg Heigold, Sylvain Gelly, et~al.
\newblock An image is worth 16x16 words: Transformers for image recognition at scale.
\newblock {\em arXiv preprint arXiv:2010.11929}, 2020.

\bibitem{how_far_agi}
Tao Feng, Chuanyang Jin, Jingyu Liu, Kunlun Zhu, Haoqin Tu, Zirui Cheng, Guanyu Lin, and Jiaxuan You.
\newblock How far are we from agi: Are llms all we need?
\newblock {\em arXiv preprint arXiv:2405.10313}, 2024.

\bibitem{ocrbench}
Ling Fu, Biao Yang, Zhebin Kuang, Jiajun Song, Yuzhe Li, Linghao Zhu, Qidi Luo, Xinyu Wang, Hao Lu, Mingxin Huang, et~al.
\newblock Ocrbench v2: An improved benchmark for evaluating large multimodal models on visual text localization and reasoning.
\newblock {\em arXiv preprint arXiv:2501.00321}, 2024.

\bibitem{gemini1.5}
Gemini.
\newblock Gemini 1.5: Unlocking multimodal understanding across millions of tokens of context.
\newblock {\em arXiv preprint arXiv:2403.05530}, 2024.

\bibitem{chinchilla}
Jordan Hoffmann, Sebastian Borgeaud, Arthur Mensch, Elena Buchatskaya, Trevor Cai, Eliza Rutherford, Diego de~Las~Casas, Lisa~Anne Hendricks, Johannes Welbl, Aidan Clark, Tom Hennigan, Eric Noland, Katherine Millican, George van~den Driessche, Bogdan Damoc, Aurelia Guy, Simon Osindero, Karen Simonyan, Erich Elsen, Oriol Vinyals, Jack~W. Rae, and Laurent Sifre.
\newblock An empirical analysis of compute-optimal large language model training.
\newblock In {\em NeurIPS}, 2022.

\bibitem{4o}
Aaron Hurst, Adam Lerer, Adam~P Goucher, Adam Perelman, Aditya Ramesh, Aidan Clark, AJ~Ostrow, Akila Welihinda, Alan Hayes, Alec Radford, et~al.
\newblock Gpt-4o system card.
\newblock {\em arXiv preprint arXiv:2410.21276}, 2024.

\bibitem{o1}
Aaron Jaech, Adam Kalai, Adam Lerer, Adam Richardson, Ahmed El{-}Kishky, Aiden Low, Alec Helyar, Aleksander Madry, Alex Beutel, Alex Carney, Alex Iftimie, Alex Karpenko, Alex~Tachard Passos, Alexander Neitz, Alexander Prokofiev, Alexander Wei, Allison Tam, Ally Bennett, Ananya Kumar, Andre Saraiva, Andrea Vallone, Andrew Duberstein, Andrew Kondrich, Andrey Mishchenko, Andy Applebaum, Angela Jiang, Ashvin Nair, Barret Zoph, Behrooz Ghorbani, Ben Rossen, Benjamin Sokolowsky, Boaz Barak, Bob McGrew, Borys Minaiev, Botao Hao, Bowen Baker, Brandon Houghton, Brandon McKinzie, Brydon Eastman, Camillo Lugaresi, Cary Bassin, Cary Hudson, Chak~Ming Li, Charles de~Bourcy, Chelsea Voss, Chen Shen, Chong Zhang, Chris Koch, Chris Orsinger, Christopher Hesse, Claudia Fischer, Clive Chan, Dan Roberts, Daniel Kappler, Daniel Levy, Daniel Selsam, David Dohan, David Farhi, David Mely, David Robinson, Dimitris Tsipras, Doug Li, Dragos Oprica, Eben Freeman, Eddie Zhang, Edmund Wong, Elizabeth Proehl, Enoch Cheung, Eric Mitchell,
  Eric Wallace, Erik Ritter, Evan Mays, Fan Wang, Felipe~Petroski Such, Filippo Raso, Florencia Leoni, Foivos Tsimpourlas, Francis Song, Fred von Lohmann, Freddie Sulit, Geoff Salmon, Giambattista Parascandolo, Gildas Chabot, Grace Zhao, Greg Brockman, Guillaume Leclerc, Hadi Salman, Haiming Bao, Hao Sheng, Hart Andrin, Hessam Bagherinezhad, Hongyu Ren, Hunter Lightman, Hyung~Won Chung, Ian Kivlichan, Ian O'Connell, Ian Osband, Ignasi~Clavera Gilaberte, and Ilge Akkaya.
\newblock Openai o1 system card.
\newblock {\em CoRR}, abs/2412.16720, 2024.

\bibitem{openvla}
Moo~Jin Kim, Karl Pertsch, Siddharth Karamcheti, Ted Xiao, Ashwin Balakrishna, Suraj Nair, Rafael Rafailov, Ethan~Paul Foster, Pannag~R. Sanketi, Quan Vuong, Thomas Kollar, Benjamin Burchfiel, Russ Tedrake, Dorsa Sadigh, Sergey Levine, Percy Liang, and Chelsea Finn.
\newblock Openvla: An open-source vision-language-action model.
\newblock In Pulkit Agrawal, Oliver Kroemer, and Wolfram Burgard, editors, {\em Conference on Robot Learning, 6-9 November 2024, Munich, Germany}, volume 270 of {\em Proceedings of Machine Learning Research}, pages 2679--2713. {PMLR}, 2024.

\bibitem{alpaca_eval}
Xuechen Li, Tianyi Zhang, Yann Dubois, Rohan Taori, Ishaan Gulrajani, Carlos Guestrin, Percy Liang, and Tatsunori~B. Hashimoto.
\newblock Alpacaeval: An automatic evaluator of instruction-following models.
\newblock \url{https://github.com/tatsu-lab/alpaca_eval}, 5 2023.

\bibitem{deepseek-v3}
Aixin Liu, Bei Feng, Bing Xue, Bingxuan Wang, Bochao Wu, Chengda Lu, Chenggang Zhao, Chengqi Deng, Chenyu Zhang, Chong Ruan, et~al.
\newblock Deepseek-v3 technical report.
\newblock {\em arXiv preprint arXiv:2412.19437}, 2024.

\bibitem{llava}
Haotian Liu, Chunyuan Li, Qingyang Wu, and Yong~Jae Lee.
\newblock Visual instruction tuning.
\newblock In Alice Oh, Tristan Naumann, Amir Globerson, Kate Saenko, Moritz Hardt, and Sergey Levine, editors, {\em Advances in Neural Information Processing Systems 36: Annual Conference on Neural Information Processing Systems 2023, NeurIPS 2023, New Orleans, LA, USA, December 10 - 16, 2023}, 2023.

\bibitem{embodied}
Yang Liu, Weixing Chen, Yongjie Bai, Xiaodan Liang, Guanbin Li, Wen Gao, and Liang Lin.
\newblock Aligning cyber space with physical world: A comprehensive survey on embodied ai.
\newblock {\em arXiv preprint arXiv:2407.06886}, 2024.

\bibitem{llama3}
Meta.
\newblock Introducing meta llama 3: The most capable openly available llm to date.
\newblock \url{https://ai.meta.com/blog/meta-llama-3/}, 2024.

\bibitem{llamaquestions}
Eliya Nachmani, Alon Levkovitch, Roy Hirsch, Julian Salazar, Chulayuth Asawaroengchai, Soroosh Mariooryad, Ehud Rivlin, RJ~Skerry-Ryan, and Michelle~Tadmor Ramanovich.
\newblock Spoken question answering and speech continuation using spectrogram-powered llm.
\newblock {\em arXiv preprint arXiv:2305.15255}, 2023.

\bibitem{GPT-4}
OpenAI.
\newblock {GPT-4} technical report.
\newblock {\em CoRR}, abs/2303.08774, 2023.

\bibitem{librispeech}
Vassil Panayotov, Guoguo Chen, Daniel Povey, and Sanjeev Khudanpur.
\newblock Librispeech: an asr corpus based on public domain audio books.
\newblock In {\em Acoustics, Speech and Signal Processing (ICASSP), 2015 IEEE International Conference on}, pages 5206--5210. IEEE, 2015.

\bibitem{whisper}
Alec Radford, Jong~Wook Kim, Tao Xu, Greg Brockman, Christine McLeavey, and Ilya Sutskever.
\newblock Robust speech recognition via large-scale weak supervision.
\newblock In {\em International conference on machine learning}, pages 28492--28518. PMLR, 2023.

\bibitem{cambrian}
Peter Tong, Ellis Brown, Penghao Wu, Sanghyun Woo, Adithya Jairam~Vedagiri IYER, Sai~Charitha Akula, Shusheng Yang, Jihan Yang, Manoj Middepogu, Ziteng Wang, et~al.
\newblock Cambrian-1: A fully open, vision-centric exploration of multimodal llms.
\newblock {\em Advances in Neural Information Processing Systems}, 37:87310--87356, 2024.

\bibitem{vaswani2017attention}
Ashish Vaswani, Noam Shazeer, Niki Parmar, Jakob Uszkoreit, Llion Jones, Aidan~N Gomez, {\L}ukasz Kaiser, and Illia Polosukhin.
\newblock Attention is all you need.
\newblock {\em Advances in neural information processing systems}, 30, 2017.

\bibitem{full-dup}
Peng Wang, Songshuo Lu, Yaohua Tang, Sijie Yan, Yuanjun Xiong, and Wei Xia.
\newblock A full-duplex speech dialogue scheme based on large language models.
\newblock {\em CoRR}, abs/2405.19487, 2024.

\bibitem{towardembodied}
Yequan Wang and Aixin Sun.
\newblock Toward embodied agi: A review of embodied ai and the road ahead.
\newblock {\em arXiv preprint arXiv:2505.14235}, 2025.

\bibitem{realworldqa}
X.AI.
\newblock Realworldqa, 2024.

\bibitem{mini-omni}
Zhifei Xie and Changqiao Wu.
\newblock Mini-omni: Language models can hear, talk while thinking in streaming.
\newblock {\em arXiv preprint arXiv:2408.16725}, 2024.

\bibitem{qwen-omni}
Jin Xu, Zhifang Guo, Jinzheng He, Hangrui Hu, Ting He, Shuai Bai, Keqin Chen, Jialin Wang, Yang Fan, Kai Dang, et~al.
\newblock Qwen2. 5-omni technical report.
\newblock {\em arXiv preprint arXiv:2503.20215}, 2025.

\bibitem{qwen3}
An~Yang, Anfeng Li, Baosong Yang, Beichen Zhang, Binyuan Hui, Bo~Zheng, Bowen Yu, Chang Gao, Chengen Huang, Chenxu Lv, et~al.
\newblock Qwen3 technical report.
\newblock {\em arXiv preprint arXiv:2505.09388}, 2025.

\bibitem{mu-scaling}
Yiqun Yao and Yequan Wang.
\newblock Research without re-search: Maximal update parametrization yields accurate loss prediction across scales.
\newblock {\em CoRR}, abs/2304.06875, 2023.

\bibitem{minicpm-o}
Chongyi~Wang Yuan~Yao, Tianyu~Yu et~al.
\newblock Minicpm-o 2.6: A gpt-4o level mllm for vision, speech, and multimodal live streaming on your phone, 2025.
\newblock https://openbmb.notion.site/MiniCPM-o-2-6-A-GPT-4o-Level \\-MLLM-for-Vision-Speech-and-Multimodal-Live-Streaming-on-Your-Phone.

\bibitem{glm-voice}
Aohan Zeng, Zhengxiao Du, Mingdao Liu, Lei Zhang, Shengmin Jiang, Yuxiao Dong, and Jie Tang.
\newblock Scaling speech-text pre-training with synthetic interleaved data.
\newblock {\em CoRR}, abs/2411.17607, 2024.

\bibitem{wenetspeech}
Binbin Zhang, Hang Lv, Pengcheng Guo, Qijie Shao, Chao Yang, Lei Xie, Xin Xu, Hui Bu, Xiaoyu Chen, Chenchen Zeng, et~al.
\newblock Wenetspeech: A 10000+ hours multi-domain mandarin corpus for speech recognition.
\newblock In {\em ICASSP 2022-2022 IEEE International Conference on Acoustics, Speech and Signal Processing (ICASSP)}, pages 6182--6186. IEEE, 2022.

\bibitem{omniflatten}
Qinglin Zhang, Luyao Cheng, Chong Deng, Qian Chen, Wen Wang, Siqi Zheng, Jiaqing Liu, Hai Yu, Chaohong Tan, Zhihao Du, et~al.
\newblock Omniflatten: An end-to-end gpt model for seamless voice conversation.
\newblock {\em arXiv preprint arXiv:2410.17799}, 2024.

\bibitem{beyond}
Xinrong Zhang, Yingfa Chen, Shengding Hu, Xu~Han, Zihang Xu, Yuanwei Xu, Weilin Zhao, Maosong Sun, and Zhiyuan Liu.
\newblock Beyond the turn-based game: Enabling real-time conversations with duplex models.
\newblock {\em arXiv preprint arXiv:2406.15718}, 2024.

\end{thebibliography}
